\pdfoutput=1

\documentclass[11pt]{article}

\usepackage[preprint]{acl}

\usepackage{times}
\usepackage{latexsym}

\usepackage[T1]{fontenc}

\usepackage[utf8]{inputenc}

\usepackage{microtype}

\usepackage{inconsolata}

\usepackage{graphicx}

\usepackage{multicol}

\usepackage{color}

\usepackage{color}

\usepackage{booktabs}

\newcommand{\clemgame}{\texttt{clemgame}}
\newcommand{\clembench}{\texttt{clembench}}

\title{\texttt{clembench\textsubscript{2024}}\\ A Challenging, Dynamic, Complementary, Multilingual Benchmark and Underlying Flexible Framework for LLMs as Multi-Action Agents}

\author{%
Anne Beyer, Kranti Chalamalasetti, Sherzod Hakimov\\
\textbf{Brielen Madureira, Philipp Sadler, David Schlangen${^\mathbf{1}}$ }
\thanks{$\;$ Contributions: AB designed and ran the multilingual experiments. KC updated the \texttt{wordle} games; BM did so for \texttt{private/shared}; SH did so for \texttt{drawing} and \texttt{reference}, managed the leaderboard and co-managed the project. PS maintained the main framework and the server infrastructure. DS co-designed the experiments, co-managed the project, and edited the document. All authors discussed all content.}\\
Computational Linguistics, Department of Linguistics\\
University of Potsdam, Germany\\
$^{\mathbf{1}}$German Research Center for Artificial Intelligence (DFKI), Berlin, Germany\\
\texttt{\textit{first}.\textit{last}@uni-potsdam.de}
}

\begin{document}
\maketitle

\begin{abstract}
    It
    has been established in recent work %
    that Large Language Models (LLMs)
    can be prompted to ``self-play'' conversational games that probe certain capabilities (general instruction following, strategic goal orientation, language understanding abilities), where the resulting interactive game play can be automatically scored. In this paper, we take one of the proposed frameworks for setting up such game-play environments, and further test its usefulness as an evaluation instrument, along a number of dimensions: We show that it can easily keep up with new developments while avoiding data contamination, we show that the tests implemented within it are not yet saturated (human performance is substantially higher than that of even the best models), and we show that it lends itself to investigating additional questions, such as the impact of the prompting language on performance. We believe that the approach forms a good basis for making decisions on model choice for building applied interactive systems, and perhaps ultimately setting up a closed-loop development environment of system and simulated evaluator.    
\end{abstract}

\section{Introduction}

The possibility of coaxing agentive behaviour out of large language models (LLMs) makes a vision seem to come into reach of setting up a closed-loop development cycle where the dialogue system is formulated through a description of the task that is to be reached, and evaluated through specifying a simulated user. While there is active work on both sides of this (realising task-oriented systems with LLMs, see e.g.\ \citet{hudecek-dusek-2023-large}, and evaluating with LLM-simulated users, see e.g.\ \citet{sekulic-etal-2024-reliable}), an important foundational step is to establish the validity and limitations of the LLMs-as-agents view. In 2023, a number of frameworks appeared that tackled this task through setting up dialogue ``self-play'' of LLMs on more abstract tasks with fully specifiable goals (then often called ``dialogue games'' \citep{schlangen-2023-1}). 

For this paper, we %
continued our work with the \texttt{clemgame} framework \citep{chalamalasetti-etal-2023-clembench},
and validated some of the claims that were left for future work in the original release. Specifically, we show that a) this approach can be extended to be a \textit{dynamic} benchmark, in the sense that what is being evaluated are indeed the games themselves and not specific game instances; b) it is still a \textit{challenging} benchmark, given that the scores of even the best models are considerably below human performance, which we establish for this paper for the first time; c) it is \textit{complementary} to other benchmarks, both reference-based ones (HELM by \citet{helm2022}) and preference-based ones (Chatbot arena by \citet{chatbotarena-2024}); d) the underlying abstractions are \textit{flexible}, so that new models can be integrated easily, making it possible, as we do here, to track the rise of open-weight models since the first release of the benchmark; and, last but not least, e) the framework makes evaluation of \textit{multilingual} capabilities of models easily possible, as we exemplify here for one of the dialogue games.

Altogether, we draw from this the conclusion that \texttt{clembench} constitutes a valuable tool for the community for testing chat-optimised LLMs (and basing decisions on the outcome), but also as an instrument for detailed studies of specific aspects of LLM behaviour. We end by speculating on possible future uses, for example as a learning environment specifically for interaction, and as something that brings us closer to the vision from the opening paragraph, as a build/test framework for designing improved agents.
\vspace*{-.1cm}

\section{Dialogue Games and Agent Capabilities}

From the realisation in 2022 that LLMs could be seen as generalist ``dialogue models'' (see e.g.\ \citet{andreas-2022-language}), the idea suggested itself that they could be made to simulate all sides in a conversation, and that this could be used to evaluate certain capabilities better than dataset-based evaluations.  \citet{qiao2023gameeval} implement a small number of games (20 questions-like, social deduction game) and test them on a small number of models. \citet{li2023static} also emphasise the need to go ``beyond static datasets'' and implement some interactive tasks, which however rely on scoring through a referee-model. \citet{gong2023mindagent} integrate LLMs in more clearly situated environments such as Minecraft, augmenting the models into agents with purpose-built modules. \citet{wu2024smartplay} also implement a variety of games and test a few models. \citet{Zhou2024-sotopia} focus specifically on ``social'' skills and use a game-like setting to study free-form interactions between LLM-realised agents. \citet{duan2024gtbench} finally set up a number of zero-shot games for self-play of LLMs, comparing the apparent strategies with those known to be game-theory optimal. What these works have in common (with the exception of \cite{duan2024gtbench}) is a focus on \textit{face validity}, in that the implemented games are simply postulated as being \textit{challenging}, and no attempt is made at elucidating which aspect of the underlying construct they target. 

For the current work, we %
continue our work on
one of the frameworks that was among the first to realise this idea of ``self-play for evaluation'' (preceding the work cited above), and which also specifically focussed on \textit{construct validity}, the \texttt{clemgame/clembench} framework \cite{chalamalasetti-etal-2023-clembench}. We do not repeat these validity arguments here and just point the interested reader to the original publication; what we do here is to introduce the basic components insofar as they are relevant for the work presented here.

The main idea of this framework is that games are specified through \textit{prompt templates}, which explain the game goals to the players in natural language, through \textit{response parsing rules} that define what counts as a well-formed response, and through a game-specific \textit{game flow} that defines what counts as a terminal state. A programmatic \texttt{GameMaster} then realises game play through the instantiation of the templates with specific game instances (e.g., in a guessing game, the word to guess in this round), and the turn-by-turn prompting of \textit{players} (which can be LLMs, or human players). The resulting \textit{episodes} are then scored through game specific \textit{scoring rules}. For each game, one scoring metric is determined as the \textit{main metric} (always ranging from 0 (worst) to 100 (best)). An overall score is computed by averaging this metric by game and then over games. Games where a player violates the parsing rules count as not played (to end). We track the percentage of games played to end; this allows us to separate \textit{formatting rule following} capabilities (which are important for any use of LLMs as internal components where the output needs to be of a pre-specified \textit{form}) from the strategic game play quality. The overall score is the product of the two scores. In the following, we will denote the benchmark (set of games) as \clembench.

For the experiments reported here, we introduced a generalisation layer for accessing LLMs via various routes (e.g., locally via huggingface transformers \cite{wolf-etal-2020-transformers} or via \texttt{llama.cpp}\footnote{%
    \url{https://github.com/ggerganov/llama.cpp}
}, or via various proprietor-specific APIs). This gives us the flexibility to benchmark a large selection of models, as discussed below in Section~\ref{sec:flex}, and easily integrate new ones. We also carefully went through all components described above, and in particular corrected some parsing rules and scoring rules for some games.\footnote{%
    A detailed list of changes is available in the project repositories at \url{https://github.com/clembench/.}
} Note that this makes the scores that we report not directly comparable with those previously reported. For the subset of models that was scored both in the previously reported run and in our latest one (19 models), we calculated a rank correlation (Kendall's tau, $r_\tau$) of 0.71 ($p < .05$), i.e., a strong to very strong correlation.

\begin{table*}[ht]
\scriptsize
\begin{tabular}[t]{lrl}
\toprule
models &    sc & o/g \\
\midrule
gpt-4            & 59.49 &  \$\$ \\
claude-v1.3      & 37.07 &  \$\$ \\
gpt-3.5-turbo    & 37.02 &  \$\$ \\
text-davinci-003 & 15.78 &  \$\$ \\
vicuna-13b       &  4.24 &  ow \\
oasst-12b        &  1.74 &  ow \\
koala-13b        &  1.48 &  ow \\
falcon-40b       &  0.71 &  ow \\
luminous-supreme &  0.00 &  \$\$ \\
\bottomrule
\end{tabular}

\hfill
\begin{tabular}[t]{lrl}
\toprule
models &    sc & o/g \\
\midrule
gpt-4-0613                  & 60.90 &  \$\$ \\
gpt-4-1106-preview          & 60.33 &  \$\$ \\
gpt-4-0314                  & 58.81 &  \$\$ \\
claude-v1.3                 & 37.64 &  \$\$ \\
claude-2.1                  & 36.38 &  \$\$ \\
claude-2                    & 33.71 &  \$\$ \\
gpt-3.5-turbo-0613          & 32.53 &  \$\$ \\
gpt-3.5-turbo-1106          & 30.45 &  \$\$ \\
openchat\_3.5                & 19.72 &  ow \\
mistral-medium              & 17.99 &  ow \\
mixtral-8x7b-instruct-v0.1  & 17.81 &  ow \\
openchat-3.5-1210           & 17.61 &  ow \\
sheep-duck-llama-2-70b-v1.1 & 17.12 &  ow \\
yi-34b-chat                 & 16.77 &  ow \\
wizardlm-70b-v1.0           & 16.70 &  ow \\
tulu-2-dpo-70b              & 15.90 &  ow \\
sus-chat-34b                & 15.64 &  ow \\
claude-instant-1.2          & 15.44 &  \$\$ \\
\bottomrule
\end{tabular}

\hfill
\begin{tabular}[t]{lrl}
\toprule
models &    sc & o/g \\
\midrule
gpt-4-turbo-2024-04-09      & 58.30 &  \$\$ \\
gpt-4-0125-preview          & 52.50 &  \$\$ \\
gpt-4-1106-preview          & 51.99 &  \$\$ \\
gpt-4-0613                  & 51.09 &  \$\$ \\
gpt-4o-2024-05-13           & 48.34 &  \$\$ \\
claude-3-opus-20240229      & 42.42 &  \$\$ \\
gemini-1.5-pro-latest       & 41.72 &  \$\$ \\
llama-3-70b-instruct        & 35.11 &  ow \\
claude-2.1                  & 32.50 &  \$\$ \\
gemini-1.5-flash-latest     & 32.00 &  \$\$ \\
claude-3-sonnet-20240229    & 30.53 &  \$\$ \\
qwen1.5-72b-chat            & 30.37 &  ow \\
mistral-large-2402          & 28.17 &  \$\$ \\
gpt-3.5-turbo-0125          & 27.22 &  \$\$ \\
gemini-1.0-pro              & 26.95 &  \$\$ \\
command-r-plus              & 24.94 &  ow \\
openchat\_3.5                & 23.64 &  ow \\
claude-3-haiku-20240307     & 22.49 &  \$\$ \\
sheep-duck-llama-2-70b-v1.1 & 21.50 &  ow \\
llama-3-8b-instruct         & 19.99 &  ow \\
openchat-3.5-1210           & 18.22 &  ow \\
wizardlm-70b-v1.0           & 17.40 &  ow \\
openchat-3.5-0106           & 17.10 &  ow \\
qwen1.5-14b-chat            & 16.80 &  ow \\
mistral-medium-2312         & 16.43 &  \$\$ \\
\bottomrule
\end{tabular}

\vspace*{-.2cm}
\caption{From left to right, results on the English \texttt{clembench} from June 2023, November 2023, May 2024. ``ow'': open weight models, ``\$\$'': gated models. The best gated model stayed constant (modulo fixes to scoring code, see text), open weight models gained substantially.}
\vspace*{-.5cm}
\label{tab:resultevo}
\end{table*}

\section{\textit{Flexible}: Performance over Time}
\label{sec:flex}

Table~\ref{tab:resultevo} shows the \clembench\ results over various timepoints, from the results of the initial publication, over an intermediate point (November 2023), to current results.\footnote{%
    The current leaderboard and all previous versions can always be found at \url{https://clembench.github.io/leaderboard.html}, with detailed result logs at \url{https://github.com/clembench/clembench-runs}.
}

Several things are notable. First, the changes described above allowed us to keep track of the rapidly evolving field. Whereas \cite{chalamalasetti-etal-2023-clembench} only reported results for 9 models, the current version now tracks 53 models. (The figure is capped at scores below 16, to save space.)
Secondly, two interesting trends are observable:\\
$\bullet$ %
While there is more competition in the field of closed weight models, as a whole, this field has not moved up. The top position is still held by a variant of GPT-4, and the top score has also not improved (insofar as the numbers are directly comparable; see discussion above). This indicates a certain \textit{saturation in achievable performance}.\\
$\bullet$ %
\textit{Open weight models}, on the other hand, \textit{have improved dramatically} over this time span. While the distance between the best open and the best gated model was 55.25 points in June 2023, it was reduced to 41.18 five months later (November 2023), and now (May 2024) stands at 24.94, thanks to the singular performance of \texttt{llama3-70b-ins}. (As the full tables in the Appendix~\ref{sec:app-results} show, this is partially due to the much improved formatting rule following capabilities of these models.)

This indicates, we believe, the usefulness of \clembench\ as an instrument for tracking developments in the field, in particular with respect to the suitability of a model to be directed to enter into goal-oriented interactions.

\section{\textit{Dynamic}: Games, not Instances}

As remarked already by \citet{chalamalasetti-etal-2023-clembench}, but not followed up on, the separation between \textit{game specification} (through templates) and \textit{game instances} makes it possible to treat the games as generative devices creating a \textit{dynamic benchmark} that can more easily evade ``data contamination'' \cite{magar-schwartz-2022-data}. For the run reported above, we created new instances for all of the games contained in \clembench. For some games, this just required sampling from an already existing pool (e.g., new target words for the \textit{wordle} game), for others, this required light manual work (e.g., selecting new target words for the \textit{taboo} game following the methodology described in the original paper; creating new target images for the \textit{image} game). As reported above, the ranking correlation between the previous run and ours is high (0.71), which we take as indication that we are indeed evaluating the game, and not the instances.  

\begin{table*}[ht!]
    \centering
    { \setlength{\tabcolsep}{4pt}
    \scriptsize
    \hspace*{-.4cm}
\begin{tabular}{llllllllll}
\toprule
\textit{\% played} &             de &            en &              it &             ja &             pt &              te &              tk &             tr &              zh \\
\midrule
GPT-4 &   100.0 (0.00) &  100.0 (0.00) &    100.0 (0.00) &   100.0 (0.00) &   100.0 (0.00) &   99.44 (-0.56) &   98.33 (-1.67) &   100.0 (0.00) &  72.78 (-27.22) \\
Claude-3 &   100.0 (0.00) &  100.0 (0.00) &    100.0 (0.00) &   100.0 (0.00) &   100.0 (0.00) &    100.0 (0.00) &    100.0 (0.00) &   100.0 (0.00) &    100.0 (0.00) \\
Llama-3-70b    &   100.0 (0.00) &  100.0 (0.00) &    100.0 (0.00) &   100.0 (0.00) &   100.0 (0.00) &   99.44 (-0.56) &    100.0 (0.00) &   100.0 (0.00) &    100.0 (0.00) \\
Llama-3-8b     &  98.33 (-1.67) &  100.0 (0.00) &    100.0 (0.00) &  98.89 (-1.11) &   100.0 (0.00) &  87.78 (-12.22) &  28.89 (-71.11) &  0.0 (-100.00) &    100.0 (0.00) \\
Command-R+         &   100.0 (0.56) &  99.44 (0.00) &    100.0 (0.56) &   100.0 (0.56) &   100.0 (0.56) &  83.89 (-15.55) &  67.78 (-31.66) &   100.0 (0.56) &    99.44 (0.00) \\
Openchat      &  98.33 (-1.67) &  100.0 (0.00) &  46.67 (-53.33) &   100.0 (0.00) &  50.0 (-50.00) &   0.0 (-100.00) &   0.0 (-100.00) &  0.0 (-100.00) &  46.67 (-53.33) \\
\bottomrule
\end{tabular}

    \vspace*{-\baselineskip}
    \hspace*{-.4cm}
\begin{tabular}{llllllllll}
\toprule
\textit{quality score} &             de &            en &              it &             ja &             pt &              te &              tk &              tr &             zh \\
\midrule
GPT-4 &  85.56 (-1.66) &  87.22 (0.00) &    89.44 (2.22) &  85.56 (-1.66) &  81.11 (-6.11) &   35.2 (-52.02) &  75.14 (-12.08) &   50.0 (-37.22) &   88.55 (1.33) \\
Claude-3 &  71.11 (-6.11) &  77.22 (0.00) &   72.22 (-5.00) &  68.89 (-8.33) &  73.33 (-3.89) &  52.22 (-25.00) &  61.11 (-16.11) &  58.89 (-18.33) &  68.89 (-8.33) \\
Llama-3-70b    &  58.33 (-4.45) &  62.78 (0.00) &    66.67 (3.89) &  56.11 (-6.67) &  60.56 (-2.22) &  45.25 (-17.53) &  36.11 (-26.67) &   45.0 (-17.78) &   68.89 (6.11) \\
Llama-3-8b     &   43.5 (-4.28) &  47.78 (0.00) &  37.78 (-10.00) &  39.33 (-8.45) &  46.11 (-1.67) &  34.18 (-13.60) &  34.62 (-13.16) &       nan (nan) &  35.0 (-12.78) \\
Command-R+         &  37.22 (-1.33) &  38.55 (0.00) &   38.33 (-0.22) &  36.11 (-2.44) &  37.22 (-1.33) &    29.8 (-8.75) &   31.15 (-7.40) &   35.56 (-2.99) &   38.55 (0.00) \\
Openchat      &   35.59 (0.03) &  35.56 (0.00) &   54.76 (19.20) &   36.11 (0.55) &  56.67 (21.11) &       nan (nan) &       nan (nan) &       nan (nan) &   40.48 (4.92) \\
\bottomrule
\end{tabular}

    }
    \caption{The \textit{reference} game in different languages. Top, ``\% played'', measuring formatting rule following, bottom, ``quality score'', measuring quality of the well-formed games. In brackets the delta compared to the original English version. \textit{GPT-4}: gpt-4-turbo-2024-04-09, \textit{Claude-3}: claude-3-opus-20240229, \textit{Openchat}: openchat-3.5-0106}
    \label{tab:multiling}
\end{table*}

\section{\textit{Challenging}: Room to Grow}

\citet{chalamalasetti-etal-2023-clembench} ``suspect'' that human performance on \clembench\ would be ``near the ceiling''. We tested this assumption. Among the authors of this paper, we created pairings so as to ensure that no player played a game where they were involved in the creation of instances. Since during the work on this paper all players developed a good understanding of all games, we consider the resulting scores to represent not \textit{average} human performance, but rather \textit{human expert performance}. We played between 10 and 15 episodes per game (leaving out \textit{wordle-clue} and \textit{wordle-critic}, as these are only variants of the main \textit{wordle} game). All games were played to end, hence `\% played' is, unsurprisingly, 100 for the human players. The resultant quality scores were: \textit{wordle}: 72, \textit{taboo}: 80.5; \textit{drawing}: 95.2; \textit{reference}: 100; leading to an average of 86.93 -- indeed considerably higher than the best result reported in Table~\ref{tab:resultevo}.

\section{\textit{Complementary}: Correlations}

To investigate how the \clembench\ measures relate to what is measured via \textit{reference-based} evaluation on the one hand, and \textit{preference-based} evaluation on the other, we computed rank correlation with HELM (v1.3.0, 2024-05-07; \citet{helm2022}) and Chatbot Arena (CA; retrieved 2024-05-16; \citet{chatbotarena-2024}), respectively. With CA, \clembench\ shares 30 models. The rankings correlate highly, with Kendall's tau at 0.65 ($r_\tau$, $p < 0.05$). With HELM, it shares 18 models. The correlation is drastically lower, at 0.39 ($r_\tau$, $p < 0.05$).
This very interesting result shows that  \clembench\ 
correlates more closely to results achieved through interaction (Chatbot Arena) --- while not actually requiring human interaction and running fully offline. (For a graphical view of the ranking relations, see the Appendix~\ref{sec:app-correlations}.)

\section{\textit{Multilingual}: A Case Study}

The separation in the \clemgame\ framework of game specification and game logic makes it possible to realise the same game in different languages, simply through translating the game templates and game parsing rules. We make use of this to probe the multilingual capabilities of a subset of the models tested above, using the \textit{reference} game as a case study. (In this game, player A is presented with three, unicode character-based, 5x5 pixel images, and tasked to describe the first one. Player B is presented with the same images, potentially in a different order, and is tasked to identify the described one. Random performance would lead to a \textit{quality score} of 33.)  We selected a set of typologically diverse languages (see Appendix~\ref{sec:app-multilingual}) and asked native speakers to translate the prompts and target expressions. Table~\ref{tab:multiling} shows the impact of playing in languages other than English. The large commercial languages hold up well when it comes to following the formatting instructions (top part of the Table), as does \texttt{llama3-70b}. All models are mostly impacted by the quality of the game play. 

We leave a more detailed analysis of these results to future work, only making the point here that this case study shows the value of \clembench\ as a promising instrument for investigating multilingual interaction instruction following capabilities.

\section{Conclusions}

In this short paper, we have assessed a recently proposed evaluation approach for LLMs that complements existing reference-based and preference-based approaches. We have shown that it possesses certain desirable properties, which promise to let it keep its relevance (because it is flexible to be adapted to new models, and its dynamic nature counteracts the danger of data contamination). The games implemented in the framework appear to sit at an interesting level: They are not particularly challenging for human players, and yet they are and remain so even for the best-performing models. In a case study, we have shown that the approach can also serve to investigate multilingual capabilities. Future work may show even further use cases, for example as a learning environment in a reinforcement learning setting, or as a development environment for more applied goal-directed dialogue systems.

\clearpage
\bibliography{anthology,latex/clemb2024}

\clearpage
\appendix

\section{Full Results}
\label{sec:app-results}

\begin{table*}[ht]
\footnotesize
\begin{tabular}{lrrrl}
\toprule
models &    sc &   \%pl &    qs & o/g \\
\midrule
gpt-4            & 59.49 & 96.06 & 61.93 &  \$\$ \\
claude-v1.3      & 37.07 & 74.76 & 49.58 &  \$\$ \\
gpt-3.5-turbo    & 37.02 & 85.86 & 43.12 &  \$\$ \\
text-davinci-003 & 15.78 & 44.50 & 35.46 &  \$\$ \\
vicuna-13b       &  4.24 & 13.58 & 31.25 &  ow \\
oasst-12b        &  1.74 & 20.85 &  8.33 &  ow \\
koala-13b        &  1.48 & 14.76 & 10.00 &  ow \\
falcon-40b       &  0.71 &  0.95 & 75.00 &  ow \\
luminous-supreme &  0.00 & 16.24 &  0.00 &  \$\$ \\
\bottomrule
\end{tabular}

\ \\
\hspace*{-.8cm}
\begin{tabular}{lrrrl}
\toprule
models &    sc &   \%pl &    qs & o/g \\
\midrule
gpt-4-0613                       & 60.90 & 97.22 & 62.64 &  \$\$ \\
gpt-4-1106-preview               & 60.33 & 97.95 & 61.59 &  \$\$ \\
gpt-4-0314                       & 58.81 & 93.79 & 62.70 &  \$\$ \\
claude-v1.3                      & 37.64 & 74.24 & 50.70 &  \$\$ \\
claude-2.1                       & 36.38 & 83.08 & 43.79 &  \$\$ \\
claude-2                         & 33.71 & 82.12 & 41.05 &  \$\$ \\
gpt-3.5-turbo-0613               & 32.53 & 91.96 & 35.37 &  \$\$ \\
gpt-3.5-turbo-1106               & 30.45 & 77.12 & 39.49 &  \$\$ \\
openchat\_3.5                     & 19.72 & 57.57 & 34.26 &  ow \\
mistral-medium                   & 17.99 & 51.11 & 35.20 &  ow \\
mixtral-8x7b-instruct-v0.1       & 17.81 & 60.49 & 29.44 &  ow \\
openchat-3.5-1210                & 17.61 & 53.18 & 33.11 &  ow \\
sheep-duck-llama-2-70b-v1.1      & 17.12 & 40.82 & 41.93 &  ow \\
yi-34b-chat                      & 16.77 & 63.76 & 26.30 &  ow \\
wizardlm-70b-v1.0                & 16.70 & 51.65 & 32.34 &  ow \\
tulu-2-dpo-70b                   & 15.90 & 54.49 & 29.18 &  ow \\
sus-chat-34b                     & 15.64 & 49.75 & 31.44 &  ow \\
claude-instant-1.2               & 15.44 & 59.61 & 25.91 &  \$\$ \\
openchat-3.5-0106                & 14.33 & 48.86 & 29.33 &  ow \\
nous-hermes-2-mixtral-8x7b-dpo   & 12.69 & 57.47 & 22.08 &  ow \\
codellama-34b-instruct-hf        & 10.34 & 23.96 & 43.15 &  ow \\
vicuna-33b-v1.3                  &  9.15 & 17.47 & 52.36 &  ow \\
wizardlm-13b-v1.2                &  7.82 & 40.49 & 19.31 &  ow \\
vicuna-13b-v1.5                  &  7.21 & 34.74 & 20.74 &  ow \\
sheep-duck-llama-2-13b           &  6.74 & 34.86 & 19.34 &  ow \\
vicuna-7b-v1.5                   &  3.46 & 12.86 & 26.91 &  ow \\
tulu-2-dpo-7b                    &  3.27 & 36.29 &  9.02 &  ow \\
command                          &  3.12 & 10.01 & 31.13 &  \$\$ \\
wizard-vicuna-13b-uncensored-hf  &  2.06 &  9.49 & 21.71 &  ow \\
llama-2-13b-chat-hf              &  1.89 &  3.43 & 55.09 &  ow \\
mistral-7b-instruct-v0.1         &  1.50 & 12.86 & 11.67 &  ow \\
llama-2-70b-chat-hf              &  1.39 &  3.79 & 36.74 &  ow \\
koala-13b-hf                     &  1.25 & 23.22 &  5.38 &  ow \\
zephyr-7b-beta                   &  1.23 &  3.95 & 31.25 &  ow \\
deepseek-llm-67b-chat            &  0.77 &  2.64 & 29.17 &  ow \\
zephyr-7b-alpha                  &  0.75 &  7.51 & 10.00 &  ow \\
llama-2-7b-chat-hf               &  0.24 &  6.05 &  4.00 &  ow \\
gpt4all-13b-snoozy               &  0.00 &  2.92 &  0.00 &  \$\$ \\
deepseek-llm-7b-chat             &  0.00 &  7.44 &  0.00 &  ow \\
oasst-sft-4-pythia-12b-epoch-3.5 &  0.00 & 14.76 &  0.00 &  ow \\
falcon-7b-instruct               &  0.00 & 14.29 &  0.00 &  ow \\
\bottomrule
\end{tabular}

\begin{tabular}{lrrrl}
\toprule
models &    sc &   \%pl &    qs & o/g \\
\midrule
gpt-4-turbo-2024-04-09         & 58.30 & 94.88 & 61.45 &  \$\$ \\
gpt-4-0125-preview             & 52.50 & 94.92 & 55.31 &  \$\$ \\
gpt-4-1106-preview             & 51.99 & 98.10 & 53.00 &  \$\$ \\
gpt-4-0613                     & 51.09 & 94.88 & 53.85 &  \$\$ \\
gpt-4o-2024-05-13              & 48.34 & 85.71 & 56.40 &  \$\$ \\
claude-3-opus-20240229         & 42.42 & 83.10 & 51.05 &  \$\$ \\
gemini-1.5-pro-latest          & 41.72 & 82.14 & 50.79 &  \$\$ \\
llama-3-70b-instruct           & 35.11 & 80.72 & 43.50 &  ow \\
claude-2.1                     & 32.50 & 82.14 & 39.57 &  \$\$ \\
gemini-1.5-flash-latest        & 32.00 & 76.14 & 42.03 &  \$\$ \\
claude-3-sonnet-20240229       & 30.53 & 85.24 & 35.82 &  \$\$ \\
qwen1.5-72b-chat               & 30.37 & 80.05 & 37.94 &  ow \\
mistral-large-2402             & 28.17 & 66.86 & 42.14 &  \$\$ \\
gpt-3.5-turbo-0125             & 27.22 & 89.67 & 30.36 &  \$\$ \\
gemini-1.0-pro                 & 26.95 & 80.14 & 33.63 &  \$\$ \\
command-r-plus                 & 24.94 & 74.90 & 33.30 &  ow \\
openchat\_3.5                   & 23.64 & 63.52 & 37.22 &  ow \\
claude-3-haiku-20240307        & 22.49 & 79.52 & 28.28 &  \$\$ \\
sheep-duck-llama-2-70b-v1.1    & 21.50 & 41.19 & 52.20 &  ow \\
llama-3-8b-instruct            & 19.99 & 76.10 & 26.27 &  ow \\
openchat-3.5-1210              & 18.22 & 51.19 & 35.60 &  ow \\
wizardlm-70b-v1.0              & 17.40 & 46.19 & 37.66 &  ow \\
openchat-3.5-0106              & 17.10 & 52.57 & 32.52 &  ow \\
qwen1.5-14b-chat               & 16.80 & 40.95 & 41.02 &  ow \\
mistral-medium-2312            & 16.43 & 49.25 & 33.36 &  \$\$ \\
qwen1.5-32b-chat               & 15.41 & 63.69 & 24.19 &  ow \\
codegemma-7b-it                & 15.30 & 51.95 & 29.45 &  ow \\
dolphin-2.5-mixtral-8x7b       & 15.10 & 46.38 & 32.55 &  ow \\
codellama-34b-instruct         & 14.35 & 33.57 & 42.76 &  ow \\
command-r                      & 14.15 & 61.67 & 22.95 &  ow \\
gemma-1.1-7b-it                & 14.14 & 49.67 & 28.46 &  ow \\
sus-chat-34b                   & 14.11 & 54.40 & 25.93 &  ow \\
mixtral-8x22b-instruct-v0.1    & 12.69 & 52.14 & 24.33 &  ow \\
tulu-2-dpo-70b                 & 12.62 & 49.76 & 25.37 &  ow \\
nous-hermes-2-mixtral-8x7b-sft & 11.95 & 39.68 & 30.12 &  ow \\
wizardlm-13b-v1.2              & 11.48 & 39.57 & 29.00 &  ow \\
vicuna-33b-v1.3                & 11.27 & 23.81 & 47.32 &  ow \\
mistral-7b-instruct-v0.2       &  9.75 & 36.91 & 26.42 &  ow \\
yi-34b-chat                    &  8.27 & 40.86 & 20.25 &  ow \\
mixtral-8x7b-instruct-v0.1     &  8.17 & 47.62 & 17.15 &  ow \\
mistral-7b-instruct-v0.1       &  8.01 & 37.14 & 21.58 &  ow \\
yi-1.5-34b-chat                &  7.67 & 52.38 & 14.65 &  ow \\
vicuna-13b-v1.5                &  7.01 & 39.52 & 17.73 &  ow \\
yi-1.5-6b-chat                 &  6.73 & 41.43 & 16.25 &  ow \\
starling-lm-7b-beta            &  6.56 & 30.89 & 21.25 &  ow \\
sheep-duck-llama-2-13b         &  5.39 & 31.90 & 16.90 &  ow \\
yi-1.5-9b-chat                 &  4.37 & 38.10 & 11.48 &  ow \\
gemma-1.1-2b-it                &  2.91 & 22.62 & 12.87 &  ow \\
qwen1.5-7b-chat                &  2.58 & 30.24 &  8.53 &  ow \\
gemma-7b-it                    &  1.82 & 17.78 & 10.23 &  ow \\
llama-2-70b-chat               &  0.81 &  7.14 & 11.31 &  ow \\
qwen1.5-0.5b-chat              &  0.12 & 25.72 &  0.48 &  ow \\
qwen1.5-1.8b-chat              &  0.00 & 15.24 &  0.00 &  ow \\
\bottomrule
\end{tabular}

\caption{In order, results on the English \texttt{clembench} from June 2023, November 2023, May 2024. ``sc'' is the clemscore, ``\%pl'' is the percentage of games played formally correctly, ``qs'' is the quality of the game play of those games; ``ow'': open weight models, ``\$\$'': gated models.
}
\label{tab:full-results}
\end{table*}

See Table~\ref{tab:full-results}.

\section{Across-Benchmark Correlations}\label{sec:app-correlations}

\begin{figure*}
    \centering
        \includegraphics[width=.9\linewidth]{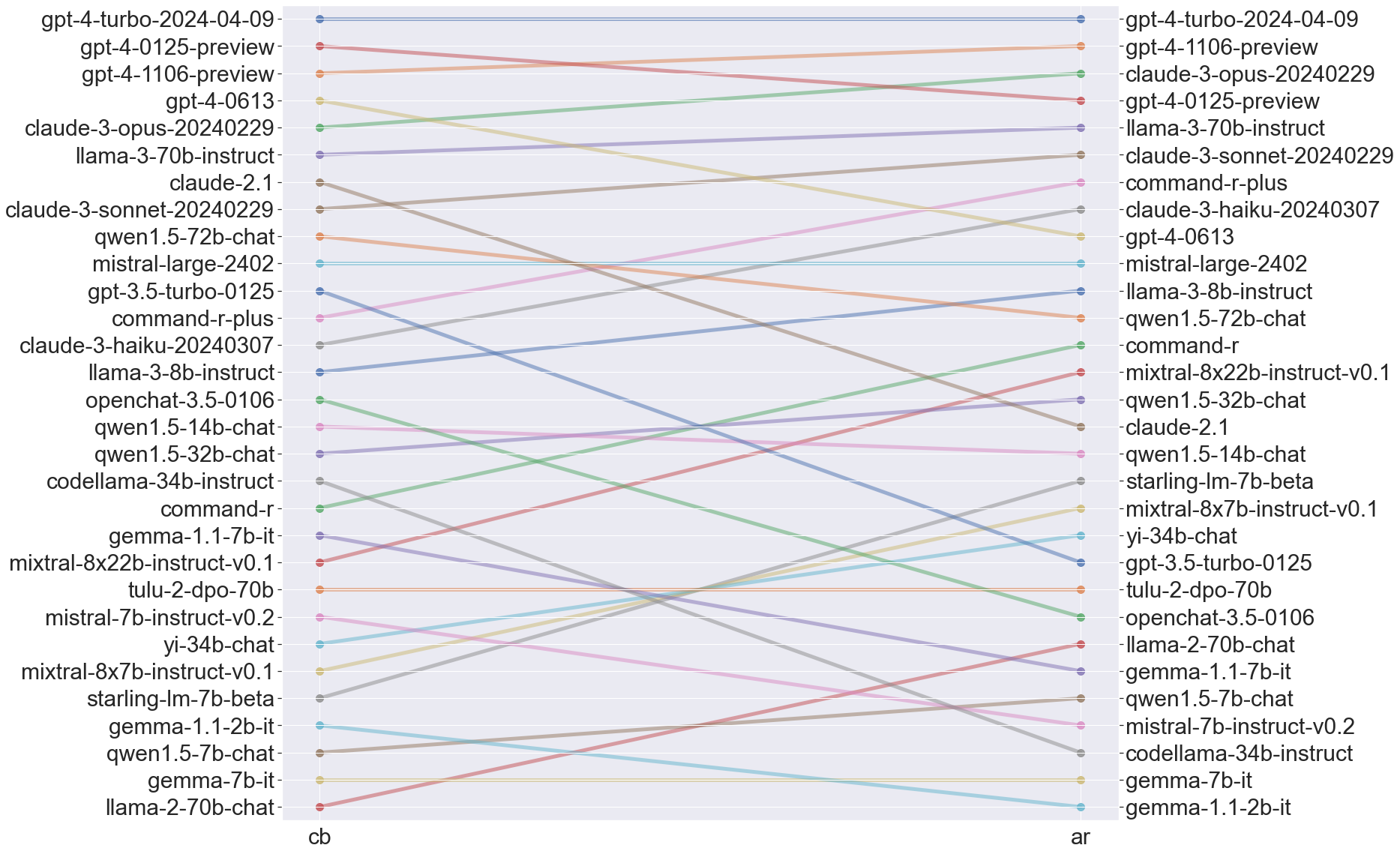}
        
        \includegraphics[width=.8\linewidth]{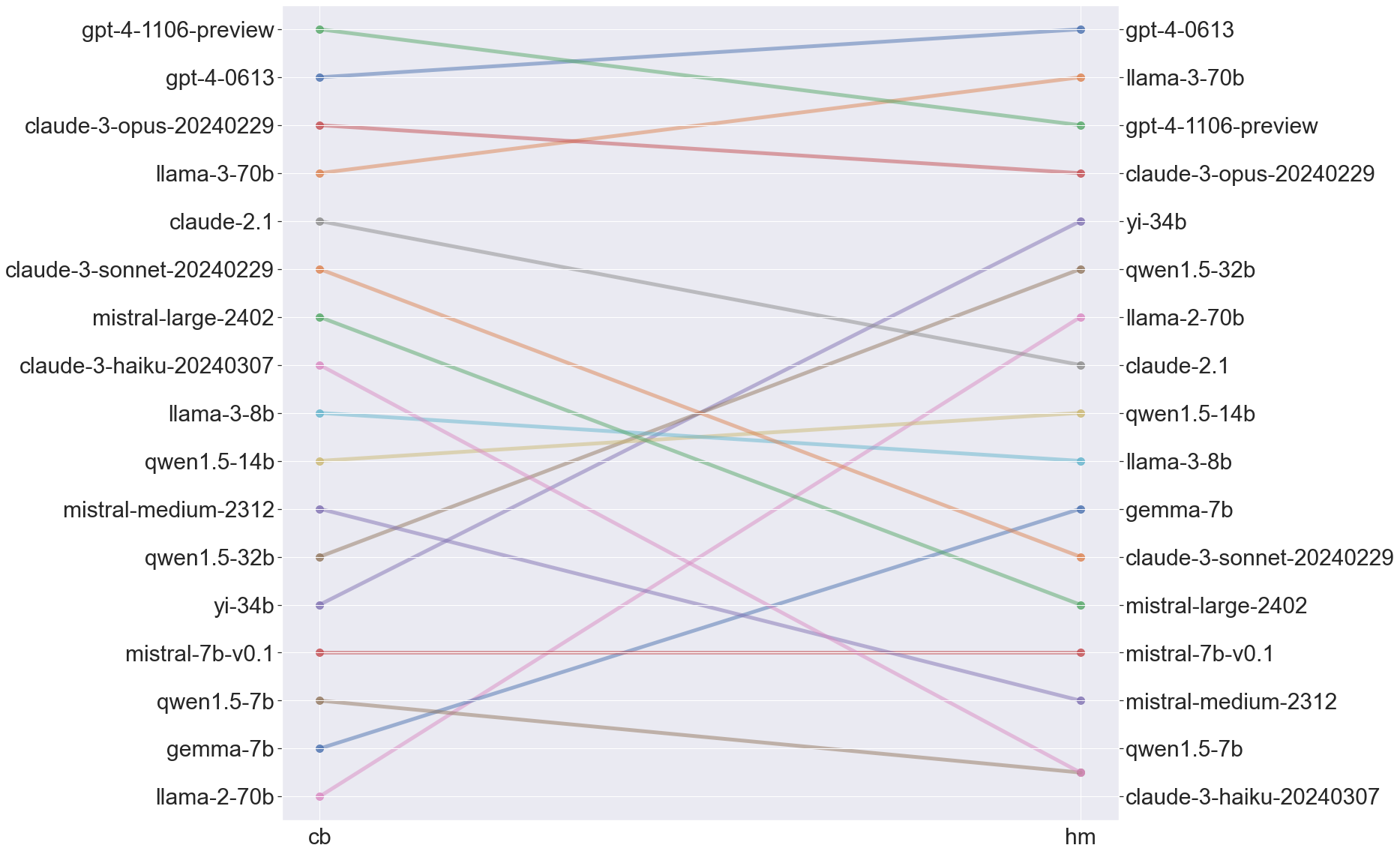}
    \caption{Top: Bump chart showing ranking differences between \clembench\ (left) and Chatbot Arena (2024-05-16; right); Bottom: Ranking differences between \clembench\ (left) and HELM (v1.3.0; right)}
    \label{fig:clemb-arena-helm}
\end{figure*}

See Figure~\ref{fig:clemb-arena-helm}.

\section{Language Tested in the Case Study}\label{sec:app-multilingual}

While there is no explicit information available on the amount of training data and optimisation for different languages for the commercial models, the model card for Command-R+ gives an overview of the supported languages on different levels.\footnote{\url{https://huggingface.co/CohereForAI/c4ai-command-r-plus}} We select a subset of eight languages from different language families, five for which the model is explicitly optimised: German (de), Italian (it), Brazilian Portuguese (pt), Japanese (ja) and Simplified Chinese (zh), one for which the training data is reported to contain resources: Turkish (tr), and two which are not supported explicitly: Telugu (te) and Turkmen (tk).

The Technical Report on GPT-4~\citet{openai2024gpt4} does not contain any information on the languages supported in the training data, but their evaluation contains a ranking of the model's performance in different languages on an automatically translated version of multiple choice questions \citep{hendrycks_measuring_2021}. Among others, our selected languages include the best and worst performing languages from their evaluation (Italian and Telugu, respectively). Similarly, the Claude-3 Model Card (available at \url{https://www-cdn.anthropic.com/de8ba9b01c9ab7cbabf5c33b80b7bbc618857627/Model_Card_Claude_3.pdf}) does not contain information on multilingual training data, but the languages we select also cover a broad range of the ones this model was evaluated on (and more).

\end{document}